\documentclass[conference]{IEEEtran}
\IEEEoverridecommandlockouts
\usepackage{cite}
\usepackage{amsmath,amssymb,amsfonts}
\usepackage{algorithmic}
\usepackage{graphicx}
\usepackage{textcomp}
\usepackage{xcolor}
\usepackage{multirow}
\usepackage{float}
\usepackage{url}
\usepackage{flushend}
\usepackage{booktabs}
\usepackage{tabularx}
\newcolumntype{Y}{>{\centering\arraybackslash}X}
\usepackage{comment}

\usepackage{framed}
\usepackage{subfig}
\usepackage{float}

\def\BibTeX{{\rm B\kern-.05em{\sc i\kern-.025em b}\kern-.08em
    T\kern-.1667em\lower.7ex\hbox{E}\kern-.125emX}}
\begin{document}

\title{Distantly-Supervised Neural Relation Extraction with Side Information using BERT  
}

\author{\IEEEauthorblockN{
Johny Moreira\,$^{1}$, 
Chaina Oliveira\,$^{1}$,
David Mac\^edo\,$^{1,2}$,
Cleber Zanchettin\,$^{1,3}$, and
Luciano Barbosa\,$^{1}$
}
\IEEEauthorblockA{\,
$^{1}$Centro de Inform\'atica, Universidade Federal de Pernambuco, Recife, Brasil\\
$^{2}$Montreal Institute for Learning Algorithms, University of Montreal, Quebec, Canada\\
$^{3}$Department of Chemical and Biological Engineering, Northwestern University, Evanston, United States of America\\
Emails: \{jms5, cso2, dlm, cz, luciano\}@cin.ufpe.br}
}

\maketitle

\begin{abstract}
Relation extraction (RE) consists in categorizing the relationship between entities in a sentence. A recent paradigm to develop relation extractors is Distant Supervision (DS), which allows the automatic creation of new datasets by taking an alignment between a text corpus and a Knowledge Base (KB). KBs can sometimes also provide additional information to the RE task. One of the methods that adopt this strategy is the RESIDE model, which proposes a distantly-supervised neural relation extraction using side information from KBs.  Considering that this method outperformed state-of-the-art baselines, in this paper, we propose a related approach to RESIDE also using additional side information, but simplifying the sentence encoding with BERT embeddings. Through experiments, we show the effectiveness of the proposed method in Google Distant Supervision and Riedel datasets concerning the BGWA and RESIDE baseline methods. Although Area Under the Curve is decreased because of unbalanced datasets, P@N results have shown that the use of BERT as sentence encoding allows superior performance to baseline methods.        
\end{abstract}


\section{Introduction}
\label{sec:introduction}

Knowledge Graphs (or Knowledge Bases) outline the relationships between entities, where the graph nodes represent the entities, and the graph edges correspond to relationships. This knowledge representation is machine-readable and is explored by different communities such as Semantic Web, which tries to improve, for instance, the results of web search engines and question-answering systems. These structures are also attractive to Artificial Intelligence researchers, which apply machine learning techniques to infer new knowledge from data \cite{nickel:2015}. However, these structures are conceptually incomplete, and there is always additional information to be added or updated. The Relation Extraction (RE) strategy is one of the techniques applied to include missing information to these bases. It extracts the semantic relationships between entities from plain texts to enhance knowledge bases. RE can be approached as a classification problem that associates the relations to categories.

To model the relation as a classification problem, we need a large labeled corpus to train supervised methods. Given the exhaustive manual effort required to build such a training set
\cite{mintz:2009} have proposed the Distant Supervision (DS) paradigm. DS aligns the structured information present in existing Knowledge Graphs to natural language texts (unstructured), assuming that any sentence containing a pair of entities (e.g.: \textbf{Barack Obama}, \textbf{United States}) from a known relation (e.g.: \textit{bornIn}) is likely to express that relation (e.g: \textbf{Barack Obama} was \textit{born in} Honolulu, \textbf{United States}.). However, this heuristic can generate noisy labeled data and degrade the extraction performance. The noisy information derives from entity pairs representing more than one relation (e.g., the entities \textbf{Barack Obama} and \textbf{United States} can also represent other relations beyond \textit{bornIn}, like \textit{employedBy}, \textit{presidentOf}, among others). The same entity pair could also appear in sentences without explicit meaningful relations (e.g., \textbf{Barack Obama} ran for the \textbf{United States} Senate in 2004).

\begin{figure*}
\centerline{\includegraphics[width=0.7\textwidth]{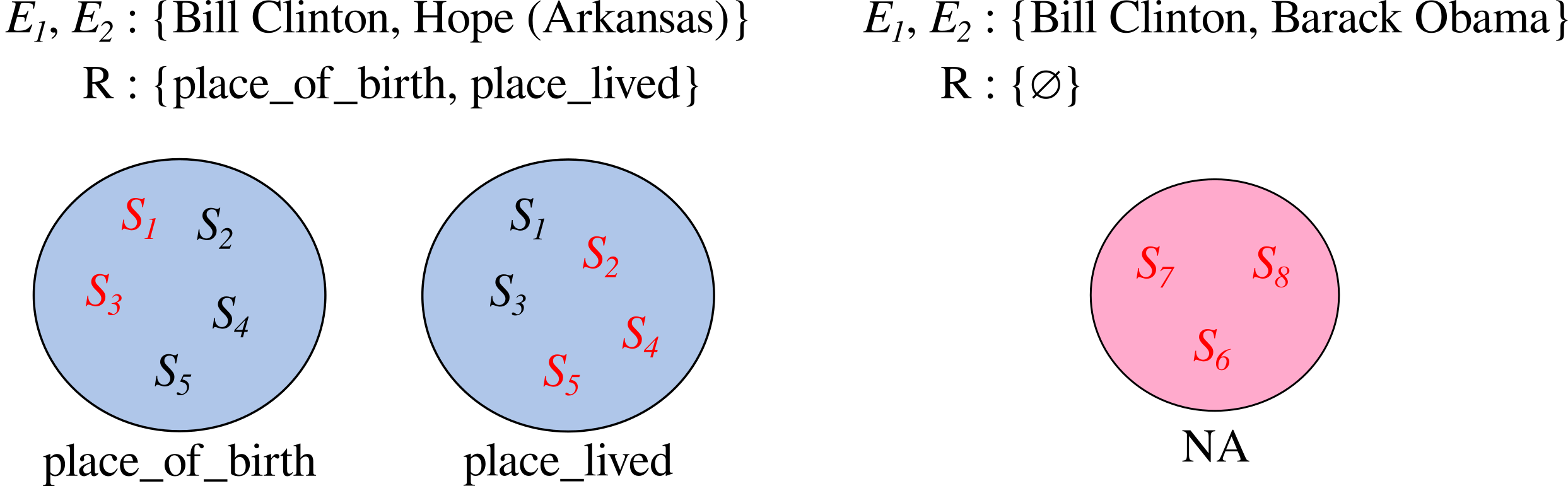}}
\caption{Relation Extraction with multi-instance learning. The blue circles represent positive bags for the given relation set (place\_of\_birth and place\_lived) for entities Bill Clinton and Hope (Arkansas). The pink circle represents a negative bag, where although the entity pair (Bill Clinton and Barack Obama) is mentioned in the sentences, no relation existing in the Knowledge Graph is expressed by them.} \label{fig:multiinstance}
\end{figure*}

The RESIDE model\footnote{https://github.com/malllabiisc/RESIDE}\cite{vashishth:2019} treats the exposed relation extraction issue as a simple classification problem with multi-instance learning, see Figure~\ref{fig:multiinstance}, where given a set of sentences $S$ and relations $R$ for an entity pair ($e_1$ and $e_2$), the task is to predict which relation $r$ holds between the two entities. The bags are samples for training the multi-instance model. A bag is composed of sentences mentioning the same pair of entities, and a single label (relation) is assigned to the entire bag. The bag is labeled as positive for a relation if at least one sentence in the bag expresses that relation. A bag is negative (not a relation) if every instance in the bag does not express any relation between the given entities. The objective of multi-instance learning is to learn separate representations for sentences and bags of sentences, trying to maximize the prediction of a set of sentences as well as single sentences.

RESIDE~\cite{vashishth:2019} presents state-of-the-art results among the relation extraction approaches. The technique uses a bidirectional Gated Recurrent Unit (bi-GRU) network to encode the local context of words in the sentence. It also applies word embeddings and tokens' position to represent the sentence. The authors also used a Graph Convolution Network (GCN) over a syntactic dependency tree to capture more substantial dependencies among words. In the last step, to sentence encoding, the authors applied an attention layer at word level. Beyond sentence encoding, it also aggregates external information, which are new embeddings for related relations and entity types.

In this paper, we propose a new distantly-supervised neural relation extraction that uses side information to RE. We use pre-trained BERT embeddings~\cite{devlin:2018}, a Bidirectional Encoder Representation for Transformers, to obtain the sentences encoding and replace the \textit{Syntactic Sentence Encoding} module of the RESIDE method (Figure \ref{fig:reside_diagram}).  Our main goal is to simplify the architecture of the efficient RESIDE method by replacing the use of word and position embeddings, Graph Convolution, Bi-GRU, and Attention mechanisms with BERT embeddings aligned with side information.

The new method called BERT-Side is evaluated and compared with state-of-the-art baselines BGWA~\cite{jat:2018}, a Bi-GRU based relation extraction model with word and sentence level attention and the RESIDE method. In the experiments, we used  Google Distant Supervision (GDS)~\cite{jat:2018} and Riedel~\cite{riedel:2010} datasets. The GDS is an extension of the Google relation extraction corpus with additional instances from entity pairs. The Riedel dataset was developed by aligning Freebase~\cite{Bollacker:2008} relations with the New York Times (NYT) corpus. 

This paper is organized as follows: Section II presents a brief background for approaches and models used in this work. Section III presents our proposed method BERT-SIDE. Section IV contains the details for the experiments applied to validate and benchmark our proposed model with baselines. Section V discusses experiments results, as well as the strongest and weakest points from the proposed approach. The paper is concluded in Section VI, showing highlights and future lines of work to improve the proposed model.

\section{Background}
\label{sec:background}

\begin{figure*}[t]
\centerline{\includegraphics[width=\textwidth]{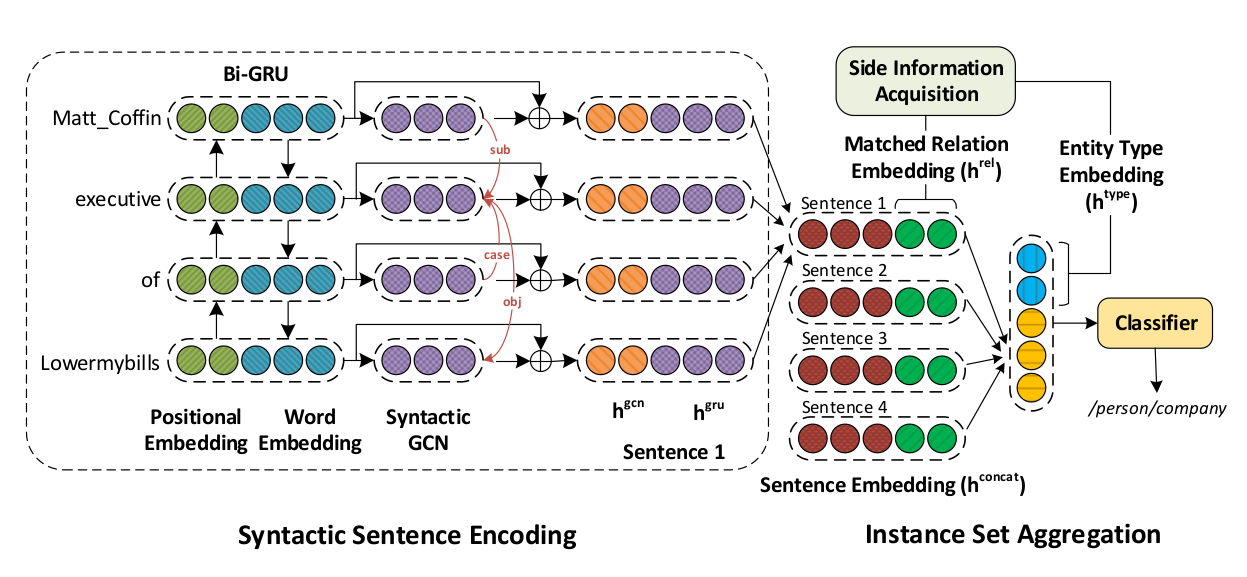}}
\caption{Overview of Reside. Source: \cite{vashishth:2019}.} \label{fig:reside_diagram}
\end{figure*}

The initial Distant Supervision paradigm does not rely on human handcrafted features, patterns, or manual annotation of training examples. Instead, it relies only on the heuristic defined by \cite{mintz:2009}. Although this approach is applicable to large and heterogeneous corpora, and also removes the manual involvement,  \cite{takamatsu:2012} highlight that this initial assumption can generate noisy labeled data and cause poor extraction performance.

Some works \cite{riedel:2010, hoffmann:2011, surdeanu:2010} have tried to enhance the DS paradigm by restricting the initial assumption; by making use of generative topic models\cite{alfonseca:2011, yao:2011, wang:2011}; and by applying probabilistic graphic models \cite{takamatsu:2012} containing hidden variables to discriminate between patterns that are expressing the relation and ambiguous ones. More recently, works have tried to leverage this problem using embedding-based methods \cite{riedel:2013}, and deep neural networks with sequence-based models or attention mechanisms \cite{lin:2016}, \cite{vashishth:2019}, \cite{jat:2018}, \cite{qu:2018}, \cite{zeng:2015}.

The model known as a pioneer on using word attention in the distant supervision context is BGWA \cite{jat:2018}. It uses a Bidirectional Gated Recurrent Unit (Bi-GRU) to encode the sentence context. The Bi-GRU runs in the two directions of the sentence (forward and backward) to capture the context of each word on both sides. These two representations are then concatenated. After this, an attention mechanism is applied over the words to obtain its importance level.

The previously mentioned RESIDE \cite{vashishth:2019} proposed a new method that uses side information in addition to the distant supervision neural method to improve the performance of relation extraction. This approach has three main modules: \textit{Syntactic Sentence Encoding}, \textit{Side Information Acquisition}, and \textit{Instance Set Aggregation} (Figure \ref{fig:reside_diagram}). The \textit{Syntactic Sentence Encoding} module utilizes a Bi-GRU over positional and word embedding to get the local context of each token. Also, a Syntactic Graph Convolution Network (GCN) is used to capture the long-range dependencies and encode this information that is further appended to the Bi-GRU output. After this, the relevance of each token is calculated using attention, and a representation of the entire sentence is obtained.

To improve the performance of the relation extraction, RESIDE makes use of outside information regards to relation alias and entity type. For alias side information acquisition it uses Open IE \cite{gabor:2015} methods to extract relevant relations between entities and align them to a paraphrase database (PPDB)~\cite{Pavlick:2015}. Further, it defines embeddings for relations and aliases. The cosine similarity is computed between them, and the closest alias relations are then matched according to a given threshold. These relation aliases representations are then concatenated to the sentence representation (output of the previous phase). In addition, for the entity types it uses the types from FIGER \cite{ling:2012} and defines an entity type embedding to each entity. If an entity has more than one type, the average is taken, and the result is also concatenated to the sentence representation.  After this, attention over sentences is applied to get the entire bag representation, and the final representation is concatenated with the embedding of the target entity types. The result is given to a softmax classifier to predict the relation. 

Proposed by \cite{devlin:2018}, BERT (Bidirectional Encoder Representation from Transformers) is a new language representation model bi-directionally trained to achieve a deeper sense of language context and flow. The authors have shown that BERT achieves state-of-the-art performance on a large number of Natural Language Processing tasks: at the sentence level, for natural language inference and paraphrasing; and at token-level, for Named Entity Recognition and Question Answering.

BERT makes use of Transformers~\cite{vaswani:2017}, which is an attention mechanism learning contextual relations between words or sub-words in a text. It is composed of an encoder, reading the text input, and a decoder producing the prediction for a given task. Once the goal of BERT is to be a language model, only the encoder is present in its architecture.

BERT is available under two model architectures. The first one is called BERT-base, and it is composed of 12 Transformers blocks, 768 hidden units, and 12 self-attention heads, totalizing up to 110M parameters. The second model, the bigger one, is composed of 27 Transformers blocks, 1024 hidden units, and 16 self-attention heads, counting 340M parameters.

These models are available as pre-trained or fine-tunned models. The pre-trained model makes use of a large set of unlabeled data for training: the BooksCorpus with 840M words, and the English Wikipedia with 2,500M words. The pre-trained model can be applied to any other task without requiring further tuning of parameters over the target task. Different from the pre-trained model, when applying the models for fine-tuning, it is required the use of labeled data to fit parameters and achieve better results for specific tasks.

BERT makes use of two training strategies for language modeling. The first one is based on a masking strategy, using a [MASK] token, where some words are hidden from the input, and the model tries to predict the masked words. The second strategy is based on Next Sentence Prediction (NSP), during input preprocessing, BERT makes use of two symbols: [CLS] as an initial marker token, and [SEP] as an indicator of sentence ending. The intuition behind the second strategy its try to predict if a second sentence is somehow connected to the first one.

Bert-as-service\footnote{\url{https://github.com/hanxiao/bert-as-service}} is a sentence encoder service making use of BERT pre-trained models. It allows the mapping of sentences into a fixed-length representation. To get the fixed-length representation, bert-as-service takes the second-to-last hidden layers of all tokens in the sentence and applies average pooling. The embedding present in [CLS] marker refers to sentence representation, while the other returned embeddings are related to the representation of each token in the sentence.

\section{BERT-Side}
\label{sec:bertside}

\begin{figure*}
\centerline{\includegraphics[width=\textwidth]{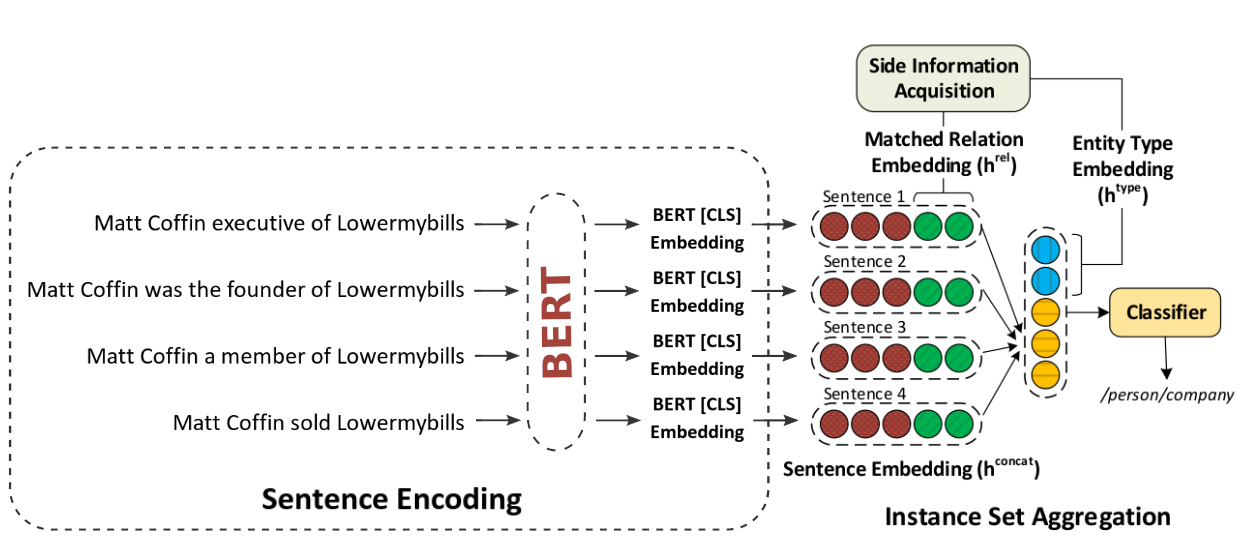}}
\caption{BERT-Side architecture. Adapted from \cite{vashishth:2019}} \label{fig:bert-side_architecture}
\end{figure*}

This section describes the proposed model BERT-Side. It is a distantly-supervised relation extraction method that uses the BERT approach (described in section \ref{sec:background}), aggregated to a side information mechanism to improve the performance of the RE task. 

According to \cite{vashishth:2019} RESIDE has outperformed state-of-the-art works. However, its \textit{Syntactic Sentence Encoding} module is rather complex. As seen in Figure \ref{fig:reside_diagram} the application of a Bi-GRU, GCN, and attention mechanisms over word and position embeddings can be simplified. BERT, as a language model, outperforms state-of-the-art tasks in Natural Language Processing, and it is a promising approach for the \textit{Syntactic Sentence Encoding} complexity of RESIDE. Hence, we replaced the \textit{Syntactic Sentence Encoding} module of RESIDE by using sentence representation obtained from BERT\cite{devlin:2018}. The new architecture is shown in Figure \ref{fig:bert-side_architecture}.

\begin{table}[t!]
{\caption{BERT-Side Models Parameters}\label{tab:model_parameters}}
\begin{tabularx}{\columnwidth}{lllcc}
\toprule
\multicolumn{2}{l}{} & \textbf{SEARCH SPACE} & \textbf{GDS} & \textbf{RIEDEL} \\
\midrule
\multirow{3}{*}{Layer 1} & Units         & {[}48, 96, 192, 384, 768{]}     & 768 & 96 \\
                         & Activation    & {[}tanh, relu, sigmoid{]} & relu & relu       \\
                         & Dropout       & Uniform(0, 1)                   & 0.58  &   0.61    \\
\midrule
\multirow{3}{*}{Layer 2} & Units         & {[}6, 12, 24, 48{]}             & 48    &  24       \\
                         & Activation    & {[}tanh, relu, sigmoid{]} & relu    &   relu    \\
                         & Dropout       & Uniform(0, 1)                   & 0.37    &  0.73   \\
\midrule
                         & Optimizer     & {[}nadam, sgd{]}            & sgd    & sgd        \\
                         & Learning rate & Uniform(0, 1)                   & 0.58 &  0.65 \\
\bottomrule
\end{tabularx}%
\end{table}

\begin{table*}
\renewcommand{\arraystretch}{1.4}
\setlength{\tabcolsep}{1pt}
\begin{center}
{\caption{Details of Riedel and GDS datasets}\label{tab:datasetsdetails}}

\begin{tabularx}{\linewidth}{c@{\hspace{1cm}}c@{\hspace{1cm}}c@{\hspace{1cm}}c}
\toprule
\textbf{Dataset}                
& \textbf{Split} & \textbf{\# Sentences} & \textbf{\# Entity-pairs} \\
\midrule
\multirow{3}{*}{\begin{tabular}[l]{@{}l@{}}\textbf{GDS} \textbf{(5 relations)}\\Jat et. al. (2018)\end{tabular}}     
& Train          & 11,297                & 6,498                    \\
& Valid          & 1,864                 & 1,082                    \\
& Test           & 5,663                 & 3,247                    \\
\midrule
\multirow{3}{*}{\begin{tabular}[l]{@{}l@{}}\textbf{Riedel (53 relations)}\\Riedel et. al. (2010)\end{tabular}} 
& Train          & 455,771                & 233,064                    \\
& Valid          & 114,317                 & 58,635                    \\
& Test           & 172,448                 & 96,678                    \\
\bottomrule
\end{tabularx}
\end{center}
\end{table*}

The sentence representation was obtained from the BERT-base model by using Bert-as-service API. The sentence embeddings come from the [CLS] marker and are obtained by average pooling over the second to the last hidden layers of the model. The sentence encoding for each sentence in a bag is concatenated with the respective alias embeddings coming from the matched relation acquired from outside sources. These representations are passed by an Attention mechanism, which is going to apply better weights to the sentences better expressing the target relation. The output of this Attention layer is the bag representation, which is then concatenated to Entity type embeddings, also coming from the information acquired from outside sources. The \textit{Side information acquisition} used by BERT-side is the same as the ones applied by the RESIDE work. The final representation is then passed through two dense, fully connected layers, each one followed by dropout. The last layer of our model is composed of a softmax activation function, which will give prediction probabilities as output. The units of the fully connected layers, dropout rates, and optimizers were sselected during training. The applied searching space and selected values for the respective hyper-parameters are shown in Table ~\ref{tab:model_parameters}. We have evaluated the model training with relation to the accuracy and using the standard cross-entropy loss for multiclass classification. 

\section{Experiments and Results}
\label{sec:experiments}

\begin{table*}[t!]
\centering
{\caption{P@N for relation extraction with different number of bags in GDS and Riedel datasets}\label{tab:p@n_gds}}
\begin{tabularx}{\textwidth}{l|YYY|YYY|YYY}
\toprule 
\multicolumn{9}{c}{\textbf{GDS}}         \\
\midrule
\multirow{2}{*}{} & \multicolumn{3}{c|}{One}                       & \multicolumn{3}{c|}{Two}                       & \multicolumn{3}{c}{All}             \\
                  & P@100         & P@200         & P@300          & P@100         & P@200         & P@300          & P@100      & P@200      & P@300      \\
\midrule
BGWA              & 0.97 & 0.91         & 0.88           & 0.94          & 0.92          & 0.893          & 0.98       & 0.95      & 0.95       \\
RESIDE            & 0.97 & 0.94         & 0.91          & 0.95          & 0.93          & 0.92           & \textbf{1.00} & 0.99       & 0.96       \\
BERT-Side         & \textbf{1.00}          & \textbf{0.99} & \textbf{0.95} & \textbf{1.00} & \textbf{0.99} & \textbf{0.99} & \textbf{1.00} & \textbf{1.00} & \textbf{1.00} \\
\midrule 
\multicolumn{9}{c}{\textbf{RIEDEL}}         \\
\midrule
\multirow{2}{*}{} & \multicolumn{3}{c|}{One}                       & \multicolumn{3}{c|}{Two}                       & \multicolumn{3}{c}{All}             \\
                  & P@100         & P@200         & P@300          & P@100         & P@200         & P@300          & P@100      & P@200      & P@300      \\
\midrule
BGWA              & 0.78 & 0.71         & 0.633           & 0.81          & 0.73          & 0.64          & 0.82       & 0.75      & 0.72       \\
RESIDE            & 0.80 & 0.75         & 0.69          & 0.83          & 0.73          & 0.71           & 0.84 & 0.78       & 0.76       \\
BERT-Side         & \textbf{1.00}          & \textbf{1.00} & \textbf{1.00} & \textbf{1.00} & \textbf{1.00} & \textbf{1.00} & \textbf{1.00} & \textbf{1.00} & \textbf{1.00} \\
\bottomrule
\end{tabularx}
\end{table*}

\begin{figure*}[t!]
\centering
  \subfloat[GDS dataset]{\includegraphics[width=0.48\textwidth]{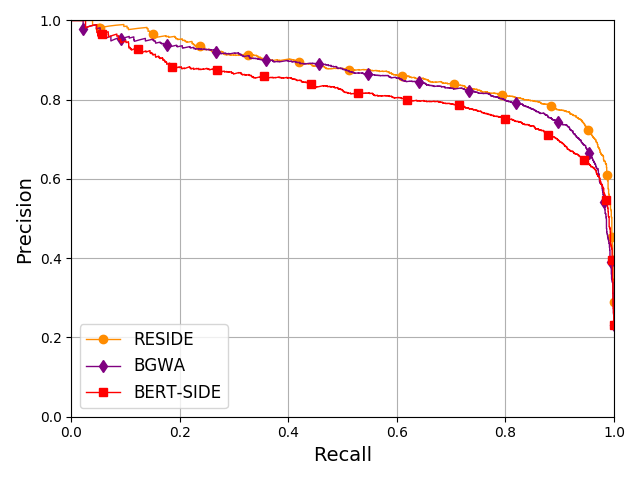}\label{fig:precision-recall-curve-gds}}
  \hfill
  \subfloat[Riedel dataset]{\includegraphics[width=0.48\textwidth]{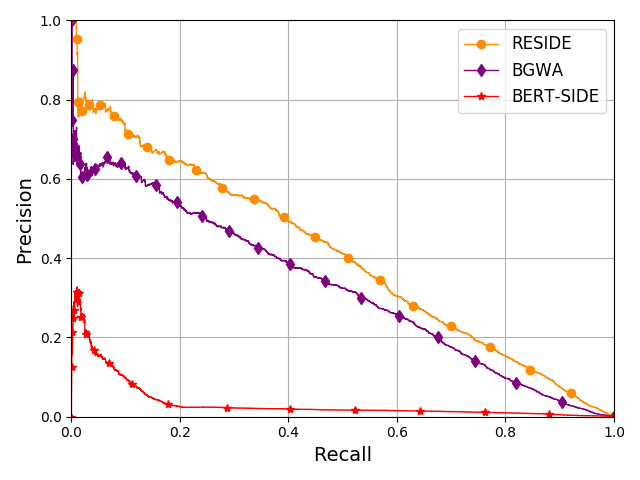}\label{fig:precision-recall-curve-riedel}}
  \hfill
\caption{Comparison between BERT-Side and baseline works on GDS and Riedel datasets}
\label{fig:precision-recall-curve}
\end{figure*}

To evaluate the BERT-Side and compare it with the state-of-the-art baselines, we used Google Distant Supervision (GDS) \cite{jat:2018} and RIEDEL \cite{riedel:2010} datasets. GDS is an extension of the Google relation extraction corpus with additional instances from entity pairs. In turn, the Riedel was built by extracting relations from the New York Times corpus using Freebase as KB. The sentences from the year 2005-2006 are used to compose the training set, while sentences from the year 2007 are used to build a test set. Some statistics details of those datasets are shown in Table \ref{tab:datasetsdetails}. In addition, BERT-Side source code is available at \url{https://github.com/guardiaum/BERT-SIDE}.

\subsection{Baselines}
\label{subsec:baselines}

We compare the proposed model with:

\begin{itemize}
    \item BGWA \cite{jat:2018}: Bi-GRU based relation extraction model with word and sentence level attention. The model was trained with adam optimizer for two epochs with learning rate and a batch size of 0.001 and 32, respectively.
    \item RESIDE \cite{vashishth:2019}: Distantly-supervised neural relation extraction using side information. It was trained with an SGD (Stochastic Gradient Descent) optimizer for four epochs with a learning rate of 0.001 and a batch size of 32.
\end{itemize}

\subsection{Model Parameters}

Table \ref{tab:model_parameters} shows the details of the best model chosen through hyperparameters optimization according to the search space given. The model is composed of two fully connected layers with dropouts, \textit{relu} as the activation function, and SGD as the optimizer.

\begin{figure*}[t!]
\centering
  \subfloat[GDS dataset]{\includegraphics[width=\columnwidth]{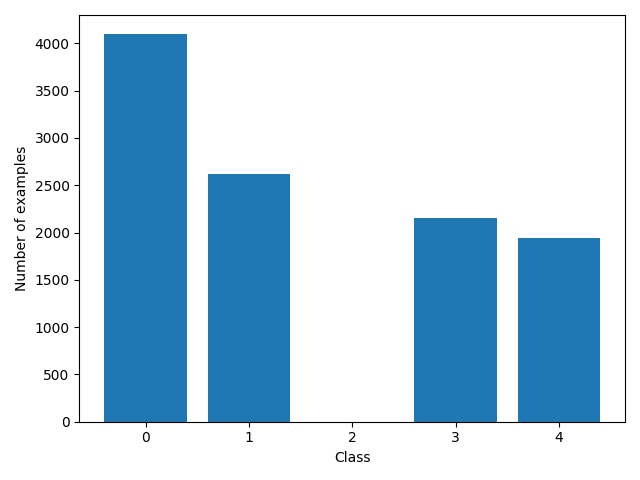}\label{fig:gds-class-dist}}
  \hfill
  \subfloat[Riedel dataset]{\includegraphics[width=\columnwidth]{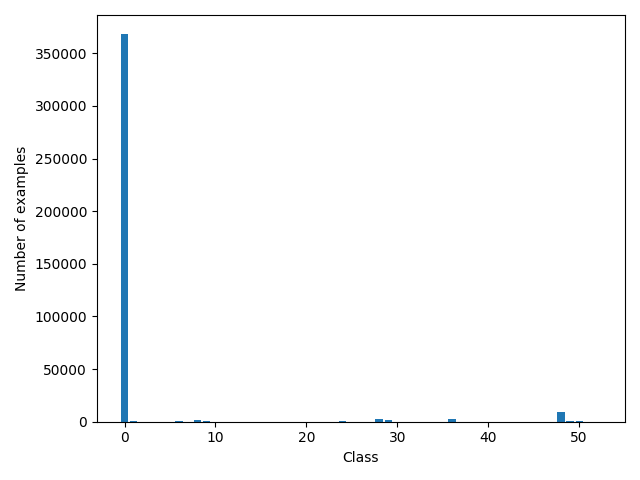}\label{fig:riedel-cls-dist}}
  \hfill
\caption{Class distribution across GDS and Riedel datasets}
\label{fig:class-dist}
\end{figure*}

The proposed work was compared with BGWA and RESIDE (described in Section \ref{sec:background}). As in \cite{vashishth:2019} we evaluate the models according to the P@N metric (the N best precision) and precision-recall curve. Table \ref{tab:p@n_gds} shows the results obtained for P@N metric when running the BERT-Side over the GDS and Riedel datasets. ``One'', ``Two'', and ``All'' represent the number of sentences randomly selected from a bag. To compute P@N for ``One'' and ``Two'' are considered only the bags with more than two sentences. The P@N metric ranks the top N sentences according to the predictions percentage. The results show that BERT-Side has outperformed baseline models when looking for random sentences selected from bags with more than two sentences (``One'' and ``two'') and also for the case of bags with one or more sentences (``All''). Statistics tests ensured this superiority of BERT-Side over baseline models. It showed that there is a statistically significant difference between P@N of BERT-Side and the other models with significance level of 0.01. Hence, BERT-Side is better in retrieving relevant sentences.

However, when looking for the Area Under the  Curve (AUC) metric, see Precision-recall curve in Figure \ref{fig:precision-recall-curve}, BERT-Side achieves results very close to baselines when looking for the GDS dataset (Figure \ref{fig:precision-recall-curve-gds}). In the case of Riedel dataset (Figure \ref{fig:precision-recall-curve-riedel}), the results are even lower. From Figure \ref{fig:class-dist}, we can notice that the distribution of samples per class is very unbalanced. Either for GDS (Figure~\ref{fig:gds-class-dist}) and Riedel (Fgure~\ref{fig:riedel-cls-dist}) datasets the class zero prevails. The class zero corresponds to the ``NA'' relation, which refers to bags with sentences expressing no relation between given entities. Hence, given the high relevance of correctly predicted sentences obtained by P@N and the low AUC results, we can assume that the unbalanced datasets have a negative impact on the BERT-Side model. As shown in Figure \ref{fig:class-dist}, the unbalancing in the Riedel dataset is greater than in the GDS dataset. Thus the negative impact was greater when running the model for the Riedel dataset.

Note that BERT-Side outperformed baseline works when calculating P@N measure, but did not outperform according to precision-recall curve. It occurs because of the P@N metric measures the relevance of the correctly classified sentences, which in the BERT-Side case always predict relevant sentences with high certainty percentage.

\section{Conclusion and Future Work}
\label{sec:conclusion}

In this paper, we proposed BERT-Side, a distantly-supervised neural relation extractor. BERT-Side uses additional KB information aligned to BERT. The experiments were executed in benchmarks datasets for relation extraction (GDS and Riedel), and the results showed that this combination has the potential to be explored. In the future, we aim to use other metrics to validate model training, e.g., AUC, as well as apply a different objective function that will handle the unbalancing in the data. %
In addition, we intend to perform fine-tuning of BERT models over these benchmark datasets, and investigate the suitability of the generated sentence representations in achieving better performance for relation extraction with the BERT-Side model.


\begin{thebibliography}{00}

\bibitem{alfonseca:2011}E. Alfonseca, K. Filippova, J. Y. Delort, and G. Garrido. "Pattern Learning for Relation Extraction with a Hierarchical Topic Model". Proceedings of the 50th Annual Meeting of the Association for Computational Linguistics: Short Papers-Volume 2, (July 2011), pp.54--59.

\bibitem{Bollacker:2008}K. Bollacker, C. Evans, P. Paritosh, T. Sturge, and J. Taylor. "Freebase: a collaboratively created graph database for structuring human knowledge". In Proceedings of the 2008 ACM SIGMOD international conference on Management of data. 2008 (pp. 1247-1250).

\bibitem{devlin:2018}J. Devlin, M. W. Chang, K. Lee, and K. Toutanova. "BERT: Pre-training of Deep Bidirectional Transformers for Language Understanding". 2018, Retrieved from http://arxiv.org/abs/1810.04805.

\bibitem{gabor:2015}G. Angeli, M. J. J. Premkumar, and C. D. Manning. "Leveraging linguistic structure for open domain information extraction". Proceedings of the 53rd Annual Meeting of the Association for Computational Linguistics and the 7th International Joint Conference on Natural Language Processing (Volume 1: Long Papers), 2015, pp.344--354.

\bibitem{hoffmann:2011}R. Hoffmann, C. Zhang, X. Ling, L. Zettlemoyer, and D. S. Weld. "Knowledge-based weak supervision for information extraction of overlapping relations". ACL-HLT 2011 - Proceedings of the 49th Annual Meeting of the Association for Computational Linguistics: Human Language Technologies, pp. 541--550. https://doi.org/978-1-932432-87-9.

\bibitem{jat:2018}S. Jat, S. Khandelwal, and P. Talukdar (2018). ``Improving Distantly Supervised Relation Extraction using Word and Entity Based Attention``. http://arxiv.org/abs/1804.06987.

\bibitem{lin:2016}Y. Lin, S. Shen, Z. Liu, H. Luan, and M. Sun. "Neural relation extraction with selective attention over instances". 54th Annual Meeting of the Association for Computational Linguistics, 2016, pp.2124--2133.

\bibitem{ling:2012}X. Ling, and D. S. Weld. "Fine-grained entity recognition". Twenty-Sixth AAAI Conference on Artificial Intelligence, 2012.

\bibitem{mintz:2009}M. Mintz, S. Bills, R. Snow, and D. Jurafsky. "Distant supervision for relation extraction without labeled data." Proceedings of the Joint Conference of the 47th Annual Meeting of the ACL and the 4th International Joint Conference on Natural Language Processing of the AFNLP: Volume 2-Volume 2. Association for Computational Linguistics, 2009.

\bibitem{nickel:2015}M. Nickel, K. Murphy, V. Tresp, E. Gabrilovich, et al. "A review of relational machine learning for knowledge graphs." Proceedings of the IEEE 104.1, 2015, pp.11--33.

\bibitem{Pavlick:2015} E. Pavlick, P. Rastogi, J. Ganitkevitch, B. Van Durme, and C. Callison-Burch. "PPDB 2.0: Better paraphrase ranking, fine-grained entailment relations, word embeddings, and style classification". In Proceedings of the 53rd Annual Meeting of the Association for Computational Linguistics and the 7th International Joint Conference on Natural Language Processing (Volume 2: Short Papers). 2015 (pp. 425-430).

\bibitem{qu:2018}J. Qu, D. Ouyang, W. Hua, Y. Ye, and X. Li. "Distant supervision for neural relation extraction integrated with word attention and property features. Neural Networks", 2018, pp.59--69. https://doi.org/https://doi.org/10.1016/j.neunet.2018.01.006.

\bibitem{riedel:2010}S. Riedel, L. Yao, and A. McCallum. ``Modeling Relations and Their Mentions without Labeled Text``. Proceedings of the 2010 European Conference on Machine Learning and Knowledge Discovery in Databases: Part III, pp.148--163.

\bibitem{riedel:2013}S. Riedel, L. Yao, A. McCallum. and B.M. Marlin. "Relation extraction with matrix factorization and universal schemas". Proceedings of the 2013 Conference of the North American Chapter of the Association for Computational Linguistics: Human Language Technologies, June 2013 (pp. 74-84).

\bibitem{surdeanu:2010}M. Surdeanu, J. Tibshirani, R. Nallapati, and C. D. Manning. "Multi-instance Multi-label Learning for Relation Extraction". July 2010, pp.455--465.

\bibitem{takamatsu:2012}S. Takamatsu, I. Sato, and H. Nakagawa. "Reducing Wrong Labels in Distant Supervision for Relation Extraction". Jeju, Republic of Korea, July 2012, pp.721--729.

\bibitem{vashishth:2019}S. Vashishth, R. Joshi, S. S. Prayaga, C. Bhattacharyya, and P. Talukdar. "RESIDE: Improving Distantly-Supervised Neural Relation Extraction using Side Information". 2019, pp.1257--1266. https://doi.org/10.18653/v1/d18-1157.

\bibitem{vaswani:2017}A. Vaswani, N. Shazeer, N. Parmar, J. Uszkoreit, L. Jones, A.N. Gomez, L.Kaiser, and I. Polosukhin. Attention is all you need. In Advances in neural information processing systems. 2017 (pp. 5998-6008).

\bibitem{yao:2011}L. Yao, A. Haghighi, S. Riedel, and A. McCallum. "Structured relation discovery using generative models". Proceedings of the Conference on Empirical Methods in Natural Language Processing, July 2011, (pp. 1456-1466).

\bibitem{zeng:2015}D. Zeng, K. Liu, Y. Chen, and J. Zhao. "Distant Supervision for Relation Extraction via Piecewise Convolutional Neural Networks". Proceedings of the 2015 Conference on Empirical Methods in Natural Language Processing, September 2015, pp.1753--1762. https://doi.org/10.18653/v1/D15-1203.

\bibitem{wang:2011}C. Wang, J. Fan, A. Kalyanpur, and D. Gondek, "Relation extraction with relation topics". Proceedings of the Conference on Empirical Methods in Natural Language Processing, July 2011 (pp. 1426-1436).

\end{thebibliography}
\end{document}